\documentclass[sigconf]{acmart}

\usepackage{subcaption}
\AtBeginDocument{%
  \providecommand\BibTeX{{%
    \normalfont B\kern-0.5em{\scshape i\kern-0.25em b}\kern-0.8em\TeX}}}

\copyrightyear{2024}
\acmYear{2024}
\setcopyright{rightsretained}
\acmConference[GECCO '24]{Genetic and Evolutionary Computation Conference}{July 14--18, 2024}{Melbourne, VIC, Australia}
\acmBooktitle{Genetic and Evolutionary Computation Conference (GECCO '24), July 14--18, 2024, Melbourne, VIC, Australia}
\acmDOI{10.1145/3638529.3654113}
\acmISBN{979-8-4007-0494-9/24/07}

\begin{document}


\title{Structurally Flexible Neural Networks: Evolving the Building Blocks for General Agents}







\author{Joachim Winther Pedersen, Erwan Plantec, Eleni Nisioti,  Milton Montero, Sebastian Risi}
\email{ {jwin, erpl, enis, mile,  sebr}@itu.dk, }
\orcid{000-0001-7170-7108}
\affiliation{%
  \institution{IT University Copenhagen}
  \streetaddress{}
  \city{Copenhagen}
  \state{}
  \country{Denmark}
}


\begin{abstract}

Artificial neural networks used for reinforcement learning are structurally rigid, meaning that each optimized parameter of the network is tied to its specific placement in the network structure. Structural rigidity limits the ability to optimize parameters of policies across multiple environments that do not share input and output spaces. This is a consequence of the number of optimized parameters being directly dependent on the structure of the network. In this paper, we present Structurally Flexible Neural Networks (SFNNs), which consist of connected gated recurrent units (GRUs) as synaptic plasticity rules and linear layers as neurons. In contrast to earlier work, SFNNs contain several \emph{different sets} of parameterized building blocks. Here we show that SFNNs can overcome the challenging \emph{symmetry dilemma}, which refers to the problem of optimizing units with shared parameters to each express different representations during deployment. In this paper, the same SFNN can learn to solve three classic control environments that have different input/output spaces. SFFNs thus represent a step toward a more general model capable of solving several environments at once.

\end{abstract}



\maketitle

\section{Introduction}
\label{section:intro}

The introduction of deep neural networks has led to remarkable improvements when it comes to the control of artificial agents either through optimization using reinforcement learning \citep{mnih2013playing, arulkumaran2017deep} or evolutionary algorithms \citep{salimans2017evolution, stanley2019designing}. One key finding has been that more diversity in the training environments leads to policies with stronger generalization capabilities (e.g., \citep{won2019learning, kumar2021rma, kirsch2022general, risi2020increasing}). In the ideal case, we would optimize our agents across a wide array of training environments with differing features, including different input and output spaces. However, most neural networks are structurally rigid. A consequence of this inflexibility is that their architectures are tied to the specifics of a particular input and output space, which prevents typical neural networks from being optimized across domains with differing dimensions.

A key challenge on the road to achieving this goal is overcoming what we refer to as the \emph{Symmetry Dilemma}. The Symmetry Dilemma is the tension between, on the one hand, gaining desirable properties that come from inducing symmetry in the optimized parameters, and on the other hand, maintaining a structure that is capable of distributing information across its nodes. If we optimize parameters to form a symmetric structure, meaning that any parameter could be placed anywhere within a network structure without changing the functionality of the network, we gain structurally flexible properties such as invariance to permutations and changes in the sizes of the input and/or output \citep{tang2021sensory, kirsch2021introducing, kirsch2022general}. However, to distribute information within the structure, each parameter must be able to contain or express individualized information. 

As an illustrative example, a fully connected neural network where each synapse weight is the exact same value is perfectly symmetric and will be invariant to any permutations of the inputs. Obviously, this neural network is not useful for creating representations with distributed information. All neurons in any layer beyond the input layer would, of course, have the same value, as they would all get the same inputs projected through the synapses. Any attempts to break the symmetry of the network to improve its representational capabilities, for example by making the connectivity sparse, would come at the cost of the network's invariance properties. Finding the right balance between these two opposite objectives is thus of vital importance if we wish to design more generally capable agents.

Instead of parameter-sharing between single-valued synapses, recent work has proposed to achieve this balance between symmetry and individualized representations by optimizing units consisting of small recurrent neural networks and connecting copies of these throughout a larger network structure to achieve a symmetric parameterization of the structure \citep{bertens2020network,kirsch2021introducing, tang2021sensory, pedersen2022minimal}. These units can then have different hidden states that are not symmetric throughout the network. In this way, there is a detachment, where the parameters that are optimized in a symmetric manner are tasked with producing asymmetric parameters. However, as explained further in Section~\ref{enu_approach}, even with this detachment, there is a risk that all the hidden states end up becoming the same.

This paper presents a novel approach called \emph{Structurally Flexible Neural Networks} (SFNN). The main features of these networks are that they consist of sparsely connected parameterized neurons and dynamic synapses. The optimized parameters are thus those of the building blocks of a neural network, separated from the network structure. Special for SFNNs, the network contains several different sets of parameterized building blocks. These are referred to as different \emph{types} of neurons and synapses. 

The experiments in this paper showcase a single parameter set capable of solving tasks with different input and output sizes.
Starting with random connectivity, and random initial synaptic weight values, the evolved plastic synapses and neurons organize during the lifetime to achieve better performance in the given environment. 
The key to achieving a versatile learning algorithm lies in structural flexibility, which liberates the model from being tied to specific input and output spaces.
The ultimate objective is to devise a set of building blocks, that when put together to form a network, exhibit fast adaptation across various environments, regardless of their dimensionality, and the permutations of the input and output elements. Recently, the notion of foundation models has surfaced in the context of reinforcement learning \citep{yang2023foundation}. A structurally flexible model could serve as a general foundation model \citep{bommasani2021opportunities, yang2023foundation} for reinforcement learning (RL)-type tasks, even when the tasks do not share input and output dimensions.

\section{Related Work}
\label{related}

\textbf{Plasticity as Indirect Encoding.} Our approach uses Gated Recurrent Units (GRUs) \citep{chung2014empirical} with shared parameters to update synaptic weight values. As such, it is related to approaches that optimize plasticity mechanisms rather than directly optimizing weights.
Indirect encodings offer a powerful way to compress the specification of large systems - like the weights of a neural network - using smaller functions \citep{zador2019critique, pedersen2021evolving,stanley2003taxonomy}. From this perspective, approaches that optimize plasticity mechanisms rather than directly optimizing weights can be seen as a way to encode multiple realizations of the parameters of a neural network since these can take different values at different points during the lifetime of an agent \citep{najarro2020}.  

An early version of plasticity as indirect encoding evolved a single parameterized learning rule with 10 parameters to allow a feedforward network to do simple input-output associations \citep{chalmers1991evolution}. Another approach, \emph{adaptive HyperNEAT} \citep{risi2010indirectly}, has been used to exploit the geometry of a neural network to produce patterns of learning rules.  The \emph{HyperNEAT} approach has also been used for improving a controller's ability to control robot morphologies outside of what was experienced during training \citep{risi2013confronting}.

\textbf{Graph Neural Networks.}  The network type presented in this paper consists of synapses that are themselves recurrent neural networks. This means that these synapses all have hidden state vectors associated with them, and, as explained in Section~\ref{subsection:propagation}, these are updated using local information as well as a global reward signal. In this regard, they share similarities in how Graph Neural Networks (GNNs) work. GNNs are designed to analyze data represented as graphs \citep{wu2020comprehensive}. Nodes in a GNN have internal states and pass information to their neighbors through parameterized message functions that are optimized during the network's training phase \citep{zhou2020graph}. 
A neural network with parameterized neurons and synapses being GRUs can be seen as a graph where some nodes (the synapses) can only have two neighbors (neurons), whereas the "neuron" nodes can have many "synaptic" nodes as neighbors. In the networks in this paper, these different types of nodes also differ in how they propagate information and update their states. However, the idea of optimizing how nodes with internal states interact with each other is analogous to that of GNNs. GNNs have been used to control simulated robots showing that they can take advantage of the fact that robot morphology can be expressed as a graph \citep{wang2018nervenet}. In contrast, the networks presented in this paper do not make assumptions about the spatial structure of inputs or outputs of the model but are only concerned with how the units of the network can best be updated.

\section{Structurally Flexible Neural Networks}

\label{enu_approach}
This section explains the details of the proposed structural flexible neural network approach, including how the information propagates works, how synaptic strengths are updated, the architectural details, and how the property of structural flexibility is gained.

\subsection{Information Propagation and Update of Synapses}
\label{subsection:propagation}

\begin{figure*}

\begin{center}
\includegraphics[scale=.35]{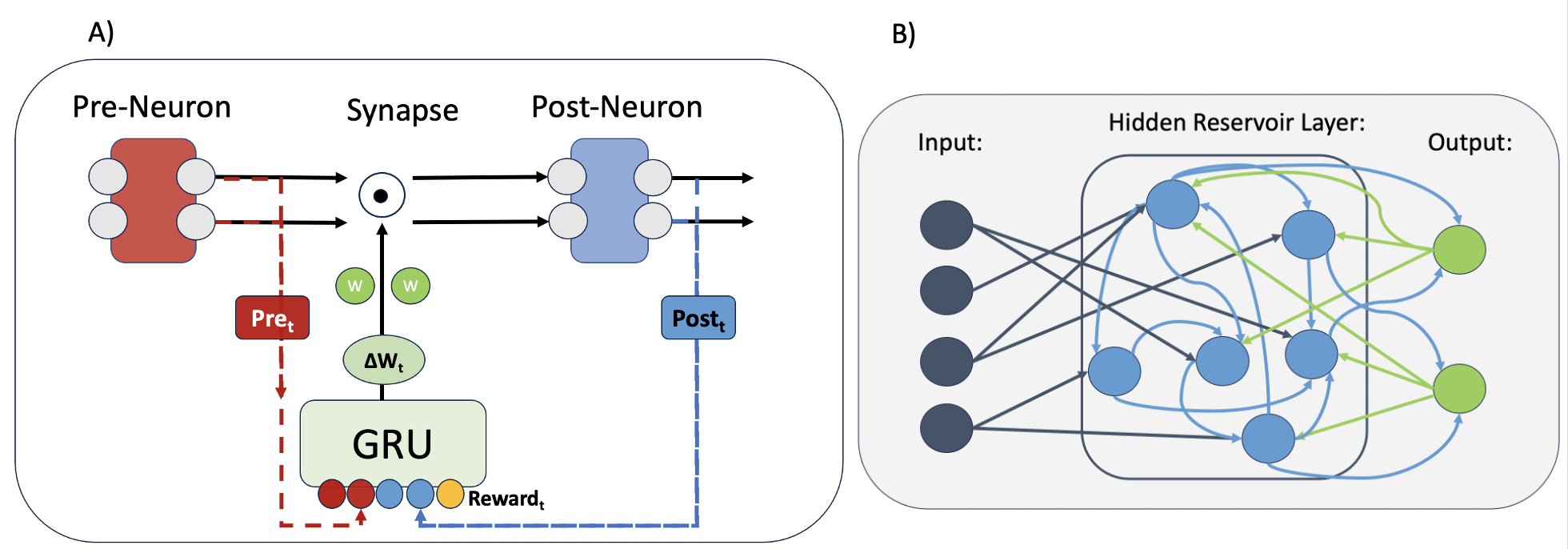}
\end{center}
\caption{\textbf{(A) Building Blocks of Structurally Flexible Neural Networks} \normalfont Depicted is a representation of the flow of activity from one neuron to another through a synapse. Neural activations are small vectors, rather than just scalars. Each neuron is represented by a small linear layer with hyperbolic-tangent non-linearities, similar to \citep{pedersen2023learning}. The output of the pre-neuron is modulated by the synapse via element-wise multiplication with the current values of the synaptic weights, and the resulting signal arrives as input to the post-neuron (solid black arrows). The output of both pre- and post-neurons plus the global reward signal is then used to compute the weight updates using the GRU (dotted arrows), which is then applied to the synaptic weights. \textbf{(B) Structurally Flexible Neural Network.} \normalfont The input vector consists of the observation from the environment, which is sent through a reservoir layer with random connectivity. Neurons of a certain type are always pre-neurons to synapses of the same type. Each color corresponds to a distinct type (input, hidden, output) that shares parameters. Each neuron has a linear layer associated with it, the parameters of which are shared by all other neurons of the same color. Likewise, each colored arrow corresponds to a synapse type. Each synapse type has one set of evolved GRU gate parameters associated with it. Importantly, even though the evolved gate parameters are shared between synapses of the same type, there is a unique hidden state associated with each single synapse, allowing for individual, history-dependent updates of each synapse.}
\label{sfann:overview2}
\centering

\end{figure*}

The networks consist of parameterized neural units and synaptic update rules. A neuron is represented by a small linear layer, and the hyperbolic tangent  as activation function. Neural activations are thus small vectors rather than scalars. A synapse is represented by a vector of the same size as the neural activations. Propagation of activity from one neuron to another is done by element-wise multiplication of the activation of the pre-synaptic neuron and the synapse connecting it to the post-synaptic neuron (Figure~\ref{sfann:overview2}). Each synaptic vector is updated as being the hidden state of a GRU. The GRU takes the respective pre- and post-neurons' activations as inputs and uses these to update its hidden state. The global reward given by the environment at the previous timestep is also concatenated with the input of the GRU. The synaptic GRUs of the network proposed here differ only from regular GRUs in that newly calculated hidden states are added to the previous hidden state instead of replacing it. A learning rate that is evolved alongside the gate parameters of the GRU is scaling the magnitude of the additions to the previous hidden state. The GRUs that make up the synapses of the network can be conceptualized as non-linear history-dependent plasticity rules, responsible for updating synaptic strengths.

Figure \ref{sfann:overview2} depicts the flow of a signal from a single neuron to another. In practice, neurons receive information from multiple other neurons through their respective synapses. The input to a neuron is thus the element-wise sum of all incoming vectors.

\label{subsection:architecture}

As depicted in Figure~\ref{sfann:overview2}, all hidden neurons of the proposed network are contained in a single reservoir layer \citep{lukovsevivcius2009reservoir}. In the main experiments of this paper, all hidden neurons and their post-synapses have tied parameters. This means that the set of GRU parameters that is responsible for updating the synaptic weights coming from hidden neurons are identical. However, it is important to note that the hidden states of these parameter-sharing GRUs are not shared. This means that the synaptic weight values can be unique for each synapse. As illustrated by the coloring of Figure \ref{sfann:overview2}, output neurons share parameters with each other, but not with the hidden neurons. The same is true for the synapses that project from the output neurons back to the hidden reservoir layer. Likewise, the synapses projecting from the input to the hidden layer share GRU parameters for their updates.

The adjacency matrix of the proposed networks is sparse, with half of the possible connections set to zero. The proportion of connections set to zero is a hyperparameter that could take other values or be optimized. For every new lifetime of an agent, a new adjacency matrix is sampled with each possible connection having an equal chance of being zero or not. As illustrated in Figure~\ref{sfann:overview2}, connections can be formed from input neurons to the hidden neurons, between hidden neurons, from hidden neurons to output neurons, and from output neurons to hidden neurons. A connection that is set to zero at the beginning of the agent's lifetime will not be updated. The initial synaptic weight values for each non-zero synapse are sampled from a uniform distribution, $\mathcal{U}$(-0.1, 0.1). Starting from this random initialization, the synapses must be updated through interactions with the environment such that a functional network emerges.

At each timestep in the environment, all neurons of the network are updated multiple times through a number of micro ticks. The activation of each input neuron is set equal to an element of the observation from the environment. Since each neuron's activation is a vector, the observation element of the respective input neuron is copied to each of the values of the neural activation.
The information from the input neurons is sent through synaptic vectors to the hidden reservoir layer. The hidden neurons are updated with the signals from the input neurons as well as signals that were sent to them from hidden- or output neurons at the previous micro tick. 

Output neurons receive signals from hidden neurons at each micro tick. In all experiments below, two micro ticks were used for each environmental timestep. After the second micro tick, the activations of the output neurons are used to determine the action that the agent will carry out. The first elements of each of the output neurons' activations are combined to form an action vector that is then passed to the environment. For environments with discrete action spaces, the index of the largest value of this action vector is chosen as the action.
At each micro tick, the first element of each of the signals from the output vectors to the hidden reservoir is set to correspond to the respective element of the action vector. The first element of the signal from the output neuron with the index corresponding to the discrete action taken is set to one and to zero for all others.

Synapses are updated after the second micro tick. Activations from the previous environmental timestep are used as pre-activations to the synaptic GRUs, and the activations resulting from the second micro tick at the current environmental timestep are used as post-activations.   

\subsection{Addressing The Symmetry Dilemma}
\label{dilemma}

Symmetric parameterization means that it is possible to switch around the placement of parameters without changing the function. The challenge that comes with constructing a parameterized approach that is symmetric and thus invariant to permutations and changes to the input or output size, is that the approach must be capable of resulting in information processing that is not symmetric, i.e., where the activation values that encode the flow of information through the symmetric architecture are themselves asymmetric.

With the network proposed in Section \ref{subsection:propagation} and \ref{subsection:architecture}, we have the pieces to address the Symmetry Dilemma. The approach presented in this paper is greatly influenced by that of \citeauthor{kirsch2020meta} ~\cite{kirsch2020meta} and \citeauthor{kirsch2021introducing} ~\cite{kirsch2021introducing}, but attempts to improve upon these through the introduction of parameterized neurons, and different sets of GRU parameters for each layer as described above.

As discussed by \citeauthor{pedersen2022minimal} ~\cite{pedersen2022minimal}, the main requirement for permutation invariance is that no parameters can be optimized relative to a fixed position in the network. Along with parameter-sharing, this can be avoided with the introduction of randomness in the network during evolution. Any aspect of the network structure that, if fixed throughout evolution, would result in the evolved parameters overfitting to it must be presented with variation during the evolution phase for the parameters to be invariant to it.

An example of such variation is that the hidden states of the synaptic GRUs can be initialized randomly to form an asymmetric structure in the network and still maintain permutation- and size-invariant properties. This is because the evolutionary optimization process has no information about these random initializations, and evolution does thus not rely on any specific initialization of the hidden states.

The same logic can be applied to different aspects of randomness in the initialization of the networks. Rather than initializing a synapse with random noise, synapses can be deleted at random, making for more sparse connectivity. As long as synapses are dropped out randomly at initialization, the evolved parameters cannot overfit to a specific connectivity. This point was also noted by \citeauthor{bertens2020network}~\cite{bertens2020network}.

As mentioned in Section \ref{subsection:architecture}, these two sources of randomness (random sparse connectivity and random weight initializations) are part of the approach proposed in this paper.

\subsection{Neural and Synaptic Diversity}
\citeauthor{kirsch2020meta} ~\cite{kirsch2020meta} introduced a fully connected network where all synapses were Long-Short Term Memory cells  (LSTMs) \citep{hochreiter1997long} sharing the same parameters. Such a network is clearly symmetric in its evolved parameters, and the idea is that LSTMs that share parameters can still develop different hidden states over time, resulting in an asymmetric network structure of the parameters that change during the lifetime of the agent, what we refer to as the plastic parameters. \citeauthor{kirsch2020meta} ~\cite{kirsch2020meta} refers to this approach as \emph{Symmetric Learning Agents} (SymLA).

However, there are some downsides to such a network architecture. One downside has to do with the option of adding more trainable parameters that are subject to evolution. From the standpoint of traditional machine learning, a surprising finding in research in deep neural networks is that overparameterization can be beneficial both in terms of trainability and generalization \citep{rocks2022memorizing, hasson2020direct}. A network such as the one presented by \citeauthor{kirsch2020meta}~\cite{kirsch2020meta} and \citeauthor{kirsch2021introducing}~\cite{kirsch2021introducing}, where synapses all share LSTM parameters can only increase parameterization by increasing the size of the LSTM unit. By allowing multiple types of neurons and synapses, evolved parameters can be added by adding more types. Note, that the computational cost of propagating through the network increases with the size and number of the network's units, not with the number of different parameter sets of these units.

The second downside has to do with the symmetry dilemma: a fully connected structure of identical LSTM units is vulnerable to hidden states of the synapses converging to similar values. When the network is first initialized, the initial hidden states (the plastic parameters) of the synapses are randomly sampled, and this is what makes the network asymmetric. The approach is fully reliant on the hidden states to be able to diverge from each other and not fall into the same attractors, even though the parameters that update them are identical and inputs are the same for all synapses that are connected to the same pre-neuron. This makes it challenging for the output neurons to be independent of each other.
The risk of having units converge to the same hidden states is even bigger in deep structures of homogeneous LSTM parameters. This can be seen by considering the case where two synapses connecting the same input element to different hidden neurons become similar to each other. The more similar they get, the greater the risk that their respective post-synaptic neurons become similar in their activations. This can result in a vicious circle where these neurons propagate more similar signals forward through all their synapses, the hidden states of which now have an even greater risk of falling into the same attractor and becoming identical. This is related to the phenomenon of \emph{oversmoothing} in graph neural networks \citep{rusch2023survey}.

As noted by \citeauthor{tang2021sensory} ~\cite{tang2021sensory}, all input units must share parameters to be invariant to permutations of the input elements. However, synapses from the input to a hidden layer do not need to share parameters with synapses between other layers to adhere to the requirements for invariance. With the proposed SFNN approach of this paper, only synapses that connect the same inputs to the same outputs must share parameters.

Having multiple sets of parameters for synaptic update rules relieves the responsibility for each of them to work across potentially significantly differing contexts throughout the network, e.g., synapses from input neurons and synapses from output neurons, and this might make solutions easier to find for the evolutionary optimization process.

\section{Experiments}
\label{experiments}

In this section, we describe the experiments to test the usefulness of having distinct parameterizations of synapses and neurons of different layers that are sparsely connected. Since the main point of SFNNs is that they are not limited to a specific input- and output space, we optimize the SFNN parameters on three different environments at the same time. 

\subsection{Environments}
\label{subsection:environments}
\begin{figure}
\includegraphics[scale=0.105]{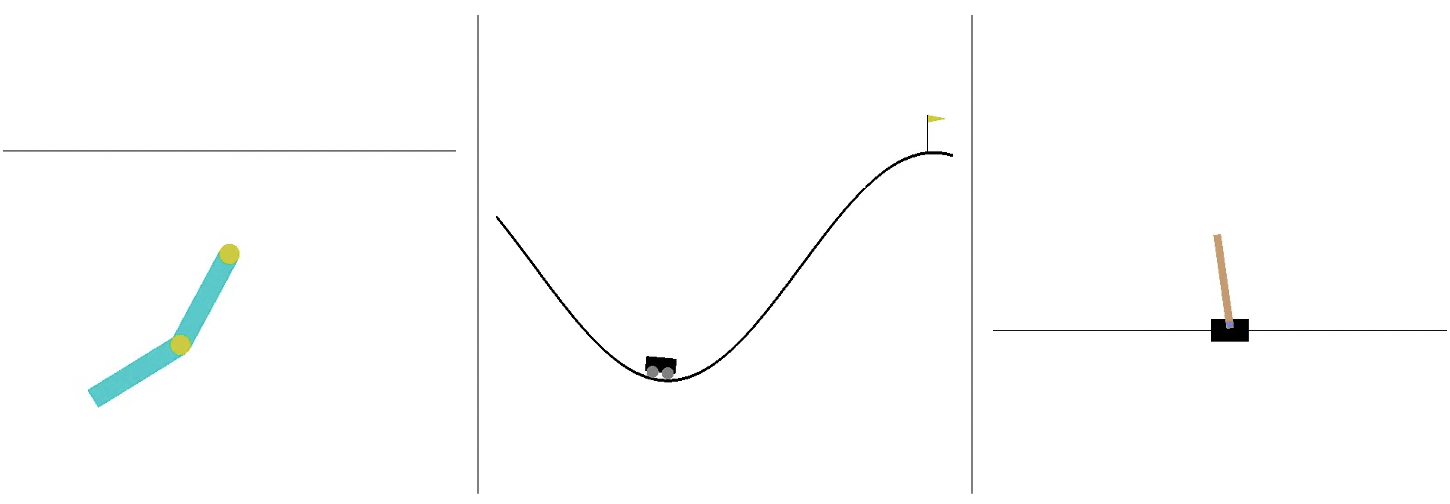}
\caption{Environments Used in Experiments \normalfont From left to right: Acrobot-v1, MountainCar-v0, CartPole-v1}
\centering
\label{fig:environments}
\end{figure}

We test our model in multiple simple control tasks (Figure~\ref{fig:environments}) implemented in the Gymnax suite \cite{gymnax2022github}. In all experiments, the ordering of the inputs to and outputs from the models stays fixed throughout the entire optimization time.

\subsubsection{CartPole-v1}
\label{subsubsection:cart}
In this task, the agent needs to control a cart on which a pole is balanced, such that the pole stays upright for as long as possible \cite{barto1983neuronlike}. The environment has four inputs and two discrete actions.

\subsubsection{Acrobot-v1}
\label{subsubsection:acro}
The task in \emph{Acrobot-v1} is to move a fixed 2-dimensional robotic arm with two joints such that the non-fixed end of the arm reaches a certain height as fast as possible. \cite{sutton1995generalization}. A reward of -1 is given for each time step spent. The environment has six inputs and three discrete actions.

\subsubsection{MountainCar-v0}
\label{subsubsection:mount}
The goal of \emph{MountainCar-v0} is to move a car from a valley on a one-dimensional track up a large hill \cite{moore1990efficient}. The car must build up momentum by first moving up a smaller hill on the opposite side. The environment has two inputs and three discrete actions. At every timestep the reward is -1, to encourage the agent to move the mountain car up the hill as fast as possible.

\subsection{Training Setup}

This section details the training setup for SFNNs in this paper. Similar to the setup of \citeauthor{kirsch2021introducing} ~\cite{kirsch2021introducing}, and \citeauthor{bertens2020network} ~\cite{bertens2020network}, the fitness of an individual is determined over a lifetime that spans multiple episodes in a given environment. However, since we are evolving parameters on three tasks, an individual actually goes through three lifetimes, one for each environment. Beyond the differences in inputs, the agent receives no information about which environment it is placed in at the beginning of a new lifetime.

When an individual is initialized for a new environment, the number of input and output units are set to fit the specifics of the particular environment. The total number of neurons in the network remains 32 in all environments. For each new lifetime, a new network is initialized with random sparse connectivity and synaptic weight values. It is then the job of the synaptic GRUs to organize the weight values in a manner that makes the network perform well in the given environment. In all experiments, the number of episodes in a lifetime is set to 8.

For each episode during the agent's lifetime, the accumulated reward for each time step is recorded. The lifetime score of the agent is then the weighted sum of recorded episode scores. Importantly, the state of the network of the agent is carried over from episode to episode, so the agent has the chance to learn throughout its lifetime. Scores obtained in episodes later in the agent's lifetime are weighted higher than ones obtained early in life. Specifically, the score of each episode is weighted by the enumeration of the episode divided by the sum of all episode enumerations. The reason for assigning a larger weight to later episodes is to encourage learning over time and to maintain performance. Agents that perform well in the first half of their lifetimes, but then start to fail, should be differentiated from agents who 
improve their scores with experience. Letting scores from the end of the agent's lifetime be more important for the lifetime score also serves to not punish early exploration.

The total fitness of an individual is its combined performance in its lifetimes in each of the three environments. Since the range from minimum to maximum score differs across the environments, the scores are mapped to be in the interval (0,1) before they are combined:
\[ f(\text{score}) = \frac{1}{{\text{max} - \text{min}}} (\text{score} - \text{min}), \] 
where \emph{min} and \emph{max} refer to the lowest and highest possible scores in the given environment. The final fitness is then the product of the three scores that are each between zero and one. Aggregating the scores with the product rather than the sum makes it necessary to score at least above zero on each of the three environments to make progress. 

The parameters of the neural and synaptic units are optimized using CMA-ES \citep{hansen2006cma, hansen2016cma}. In all experiments, the population size is set to 128. The specifics of the SFFN approach of the main experiments are summarized in Table \ref{tab:sfnn}. The linear layer of a neuron is in this case a four-by-four weight matrix. The hidden size of the GRUs is four as well, and the input to the GRU is thus two times four plus the one global reward element. For each set of GRU parameters, a single scalar is also optimized to be used as a learning rate for the hidden state updates. In practice, the learning rate is always multiplied by a constant of 0.01 to form an applied learning rate. This is done to ensure that the magnitude of the applied learning rate is smaller than the rest of the evolved parameters, at least at the beginning of the evolution phase.

\begin{table}
\begin{center}
  \caption{SFNN Hyperparameters}

  \label{tab:sfnn}
  
  \begin{tabular}{ccccccc}
   \toprule
   
    \midrule
       Neural Activation Size & 4              \\
       Synaptic Vector Size & 4                \\
       Total Number of Evolved Parameters &   565         \\ 
       Total Number of Neurons  & 32 \\
       Neural Activation Function & Tanh                  \\
       Number of Neuron Types &        3     \\
       Number of Synapse Types &        3 \\
       Number of Reservoir Ticks per Time step &       2     \\  \hline
   \bottomrule

\end{tabular}
\end{center}
\end{table}

\textbf{Variations:} We compare the SFNN approach described above with the SymLA approach implemented in the appendix of  \citeauthor{kirsch2021introducing} ~\cite{kirsch2021introducing}. We use the same hyperparameters for the LSTM sizes as in the original paper. The only difference here is that the outputs of the SymLA below were deterministic rather than sampled. In preliminary experiments, using sampled outputs led to no progress in any experiments.

We also test two ablations of the SFNN approach. In the first ablation, \emph{SFNN\_single}, all neurons and all synapses across the whole network share the evolved parameters. In the other, \emph{SFNN\_fully}, the network is fully connected, meaning that none of the possible synapses are set to zero upon initialization. The ablations share all other characteristics with the full SFNN approach.

\begin{figure}
\begin{center}
\includegraphics[scale=.175]{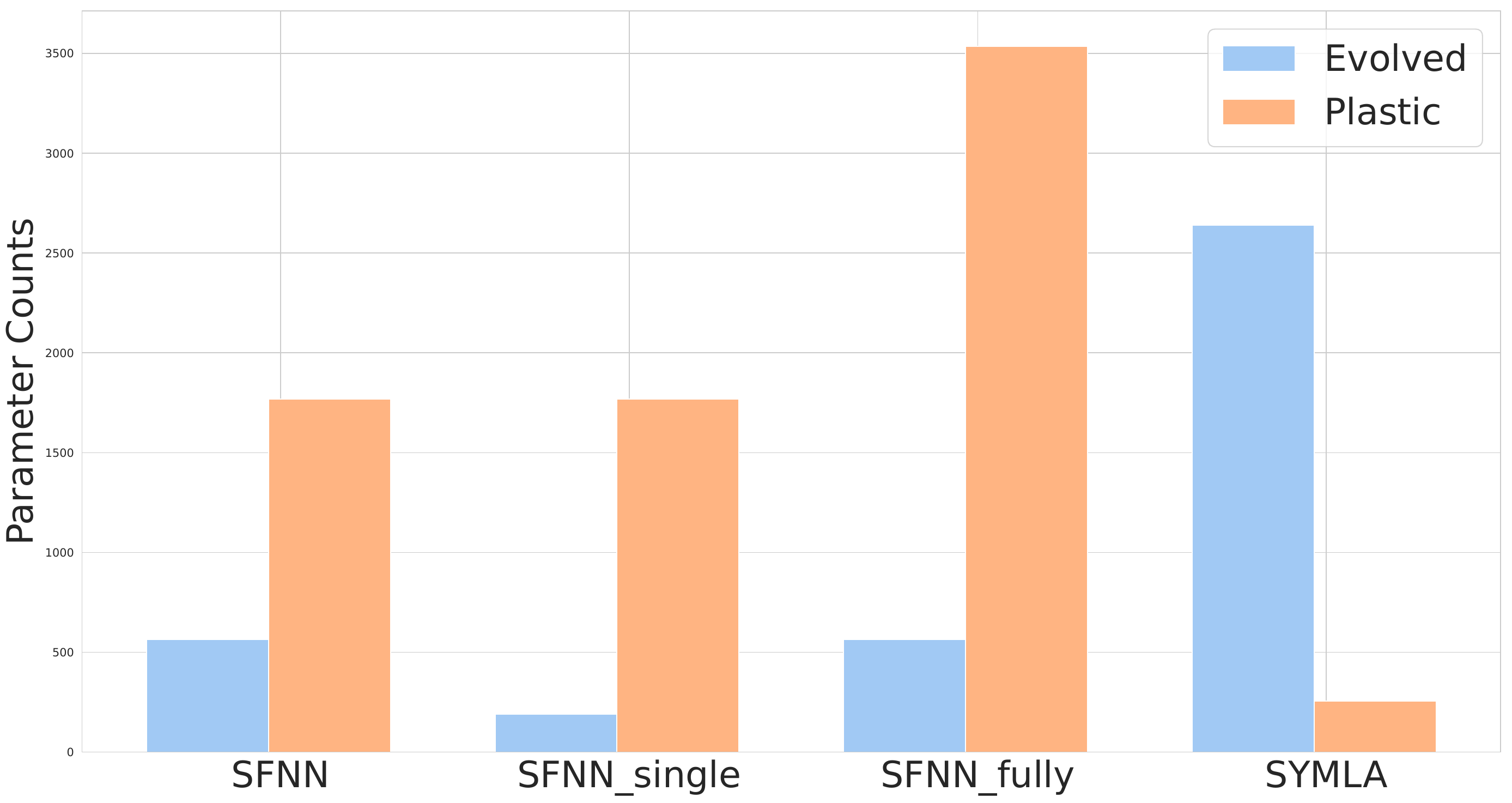}
\end{center}
\caption{\textbf{Number of evolved and plastic parameters:}  \normalfont The approaches differ in their numbers of evolved and plastic parameters. Evolved parameters are the synaptic GRU parameters and the linear layers of the neurons. The plastic parameters are the synaptic weight values  that are updated by the GRUs (LSTMs in the case of SymLA). The SFNN\_single version has the same number of plastic parameters as SFNN, but fewer evolved ones, as only parameters for a single neuron and synapse type are evolved. The SFNN\_fully version has the same number of parameters as SFNN, but has double the amount of plastic parameters, as no synapses are excluded. The SymLA method has many more evolved parameters than plastic, as the synaptic units are much bigger than in SFNN and there are fewer of them, as the SymLA networks are shallow.}
\centering
\label{fig:parameter_count}
\end{figure}

Figure~\ref{fig:parameter_count} provides an overview of how many evolved and plastic parameters each setting has.
For all settings, five evolutionary runs with different seeds were carried out.

\section{Results}
\label{sec:results}

\begin{figure*}
\begin{center}
\includegraphics[scale=.32, width=0.995\textwidth]{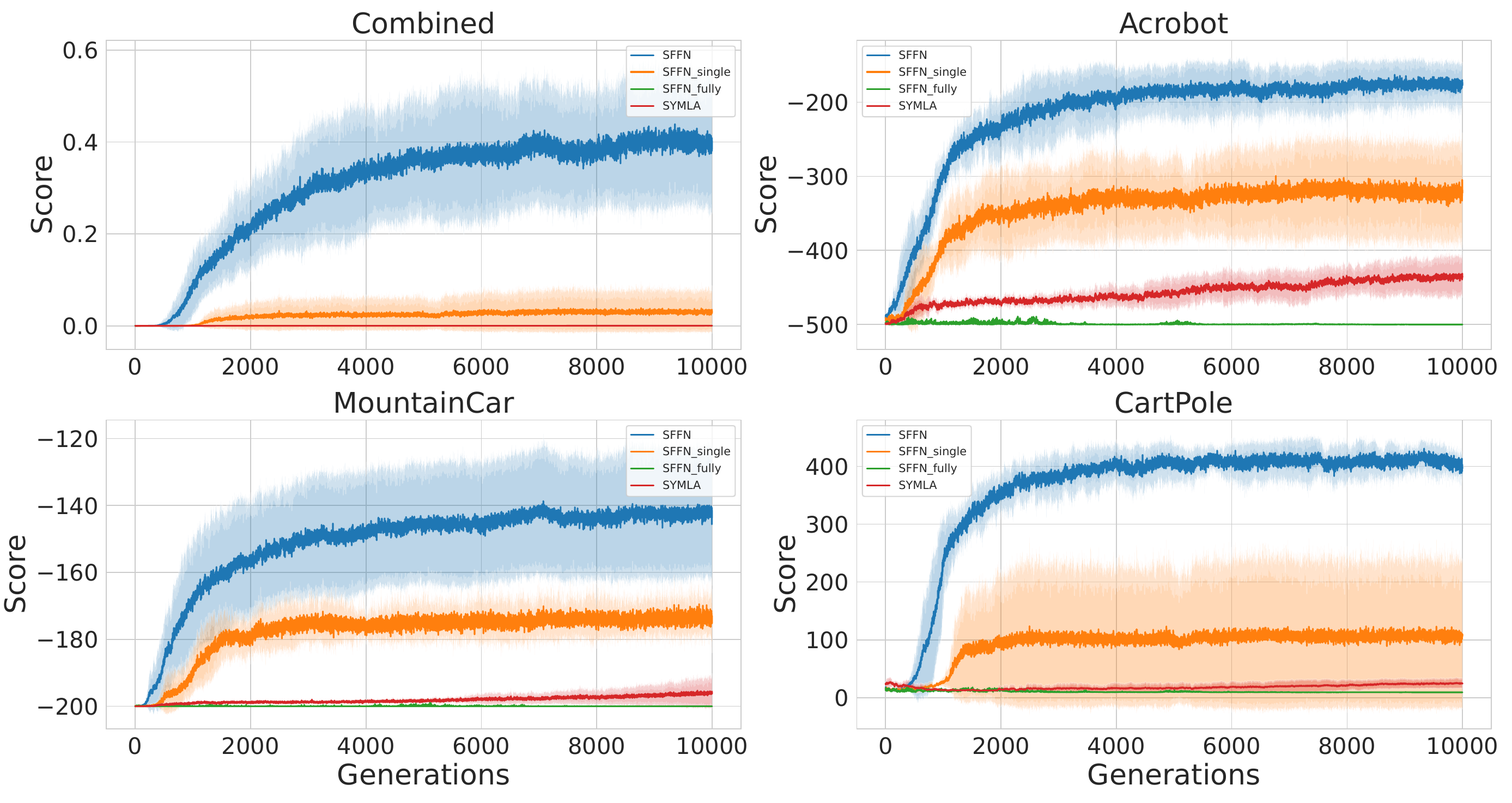}
\end{center}
\caption{\textbf{Training plots:} \normalfont Shown are the average and standard deviations of the population means of five runs for each model. Progress in each of the three environments is shown, as well as the product of each score when the scores are scaled to be between zero and one (top left). Of the four different settings, only the full SFNN can consistently make progress on each of the three environments. The fully connected SFNN shows no improvements at all, while the SFNN with a single neuron and synapse type, and the SymLA method only display modest progress.}
\centering
\label{fig:training_curves}
\end{figure*}

Training Curves of the different runs are shown in Figure \ref{fig:training_curves}. Only the SFNN approach made significant progress on all three environments at once during the evolution. 

\begin{figure}
\begin{center}
\includegraphics[scale=.25]{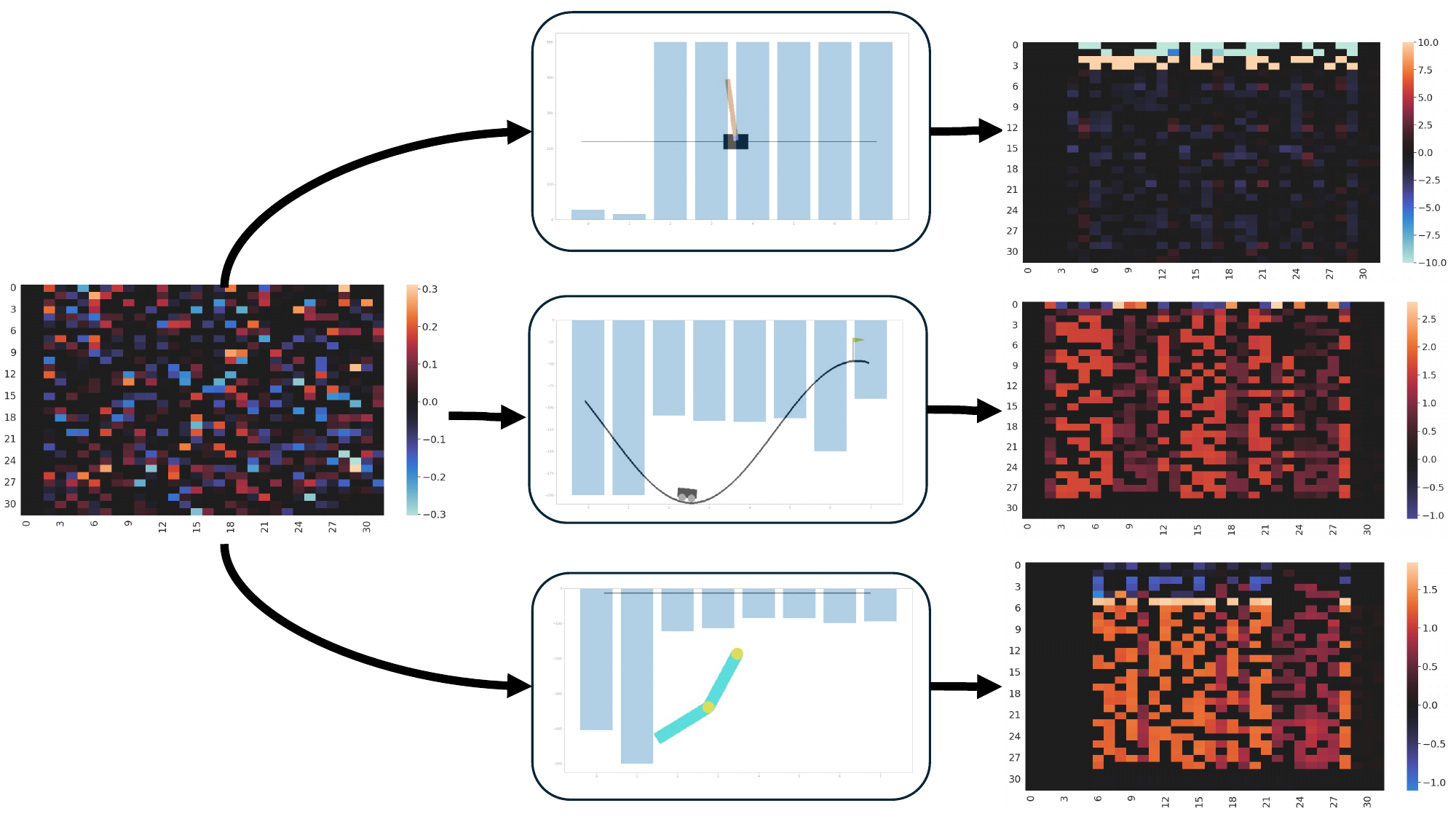}
\end{center}
\caption{\textbf{Adapting Weight Matrices in SFNNs.} \normalfont An example of the same initial weight matrix used in the three environments resulting in different weight matrices after eight episodes in a different task. The sum of the four elements that make up the synapse is depicted. Values are clipped to have a maximum magnitude of 10 for readability. The only difference in the initial matrices between the environments is how many neurons are counted as input/output neurons, to fit the respective environment. Especially the solution for the CartPole environment stands out from the two others with the magnitudes of weights coming from the input neurons being much larger than the other weights.}
\centering
\label{fig:aMats}
\end{figure}

\begin{figure*}
\begin{center}
\includegraphics[scale=.381]{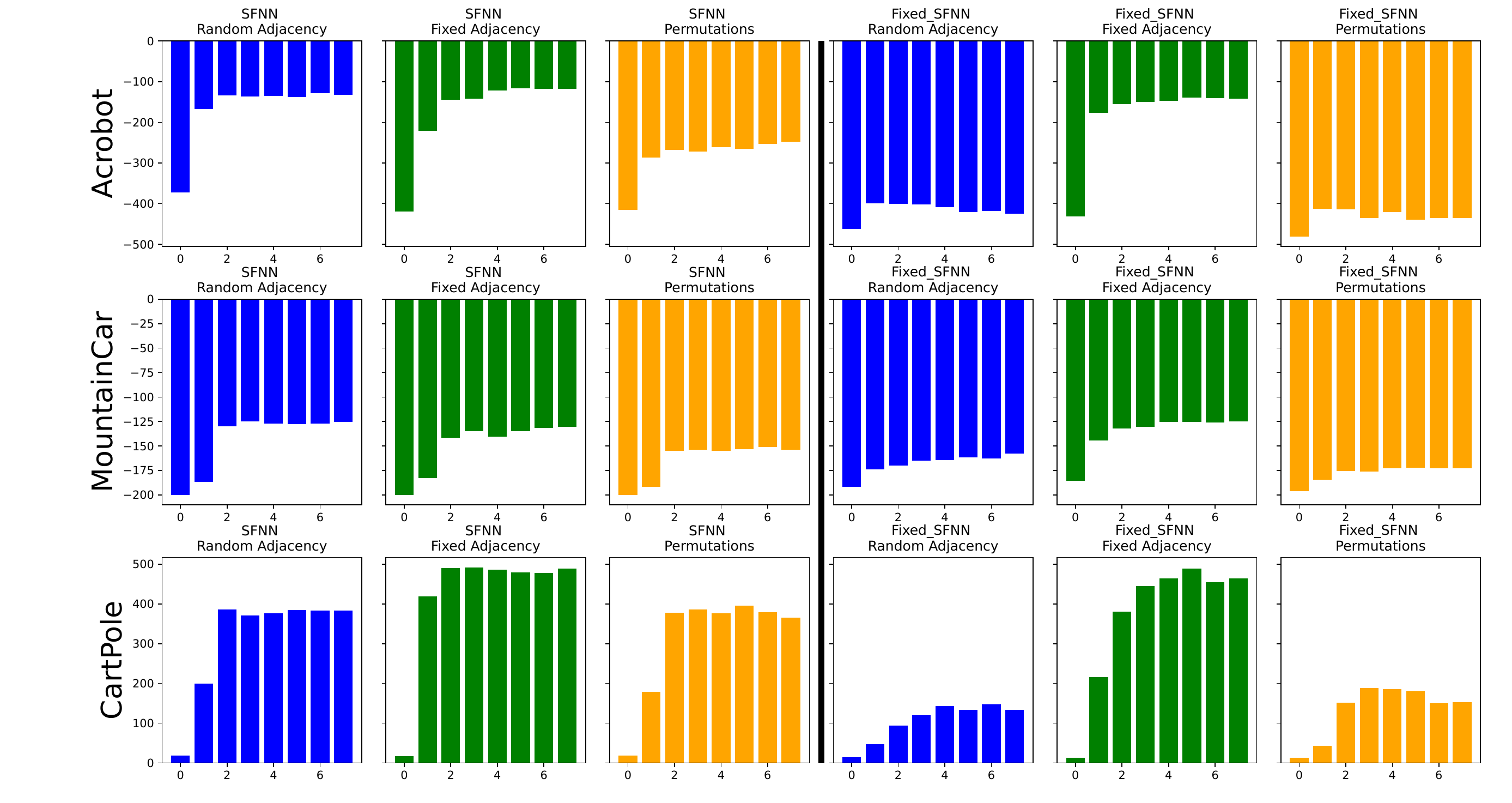}
\caption{\textbf{Lifetime Progress in Training Environments.} \normalfont Columns display the results under different conditions in each of the three environments. The shown scores are the averaged scores across 100 lifetimes. The first three columns of subplots show the results of the agent of the best of the five evolution runs of the standard SFNN approach, described in Section \ref{enu_approach}. The three remaining columns show the results of an agent trained on a single, fixed adjacency matrix throughout its evolution (Fixed\_SFNN). Each bar corresponds to the score in one episode of the lifetime of an agent. \textbf{Blue}: Adjacency matrices are initialized randomly at the beginning of the agent's lifetime. \textbf{Green}: The same adjacency is used in each lifetime. This is the same adjacency matrix used in the evolution of Fixed\_SFNN. \textbf{Yellow}: Adjacency matrices are initialized randomly and random permutations of inputs and outputs are chosen at the beginning of the agent's lifetime. The SFNN optimized under standard conditions shows similar lifetime performances across all three settings. The SFFN that was only exposed to a single adjacency matrix during its evolution obtains good scores when that specific matrix is used (green), but exhibits major performance decreases in the other two settings.
In sum, the standard SFNN approach can improve quickly across settings, while the SFNN trained on a single adjacency matrix cannot.}

\centering
\label{fig:life_progress}
\end{center}
\end{figure*}

Figure~\ref{fig:aMats} shows an example of how one initial weight matrix can diverge into different ones when the network is tasked with solving one environment or the other. This shows that the dynamics of the single set of evolved neurons and synapses can self-organize into several configurations that are adaptive to the given environment. 

Figure~\ref{fig:life_progress} shows average lifetime scores of an evolved SFNN over 100 lifetimes. Scores of the other variations from Figure \ref{fig:training_curves} are omitted, as no good solutions were found by evolution for any of them. In the first three columns, the SFNN is evaluated in the same setting as during the training, in a setting with the same adjacency matrix used in all lifetimes, and in a setting where a random permutation of the inputs and outputs are chosen at the beginning of each lifetime. In all cases, the scores of the episodes later in the agent's life tend to be better, indicating that the networks in a short amount of time are able to organize into functional networks. 

For illustration purposes, the performance of a \emph{Fixed\_SFNN} that was evolved with only a single adjacency matrix being used throughout the evolution process is also shown under the same conditions, in the three right-most columns of Figure~\ref{fig:life_progress}. While the standard SFNN is robust to input and output permutations never seen during training, the same cannot be said for the \emph{Fixed\_SFNN} which only is capable of achieving high scores when its usual adjacency matrix is used.

\section{Discussion and Future Work}
\label{discussion}

This paper introduced SFNN, a network type that has evolved parameters optimized flexibly across environments with different input and output shapes and can rapidly improve during the lifetimes of the agents that it controls. 

Of the four settings tested (Figure~\ref{fig:training_curves}), only the SFNN with different neuron and synapse types \emph{and} sparse random connectivity was able to improve on all tasks. The \emph{SymLA} approach with its relatively few plastic parameters might lack the capacity to form networks that are different enough to adapt to multiple tasks. The \emph{SFNN\_single}, on the other hand, is potentially under-parameterized in terms of the evolved parameters. \emph{SFNN\_fully} was the worst scoring setting of them all, hinting that having different neuron and synapse types alone is not enough to overcome the symmetry dilemma: a fully connected structure has a high risk of ending up with all units having the same activity, due to each unit getting the same input. 

One could imagine other types of models to train for structural flexibility and adaptation. One possibility could be a sequence-to-sequence model \citep{yousuf2021systematic} that takes input elements from the environment one at a time and outputs action elements one at a time. This model would have the same size regardless of the number of elements in the input- and output spaces. However, such a model would have some significant downsides. First, in many environments, there is no meaningful sequential information between the inputs, but the processing of each new element would be affected by the previous elements. Further, the model would still depend on seeing many permutations of input and output vectors during training to avoid overfitting to a specific ordering in a way that might be harmful to performance in novel environments. Lastly, while it is generally true for all models that an increased input- and/or output space will result in increased computational requirements, this is especially true for sequence-to-sequence models as the one imagined here, which would need an entirely new forward propagation for each new element in the input. 

Another option for more flexible neural networks could be the use of Set Transformers that are specifically designed to be permutation invariant and would not rely on being exposed to all permutations during the training phase \citep{lee2019set}. To make this type of model invariant to the length of the observation vectors from different environments, the set would be the observation elements for a given environment, instead of sequences of multiple time steps. This is similar to having parameter-sharing input neurons and synapses as with SFNNs. The attention matrix of the Transformer can be interpreted as plastic parameters \citep{schlag2021linear}. However,  whereas the size of the attention matrix of the Transformer is tied to the number of input elements, SFNNs allow the number of plastic parameters to be independent of the number of input elements.
While not part of the experiments shown in this paper, true structurally flexible neural networks should also be able to have varying hidden layer sizes across different lifetimes, enabling a trade-off between redundancy and stability on one hand, and computational efficiency on the other.  

The SFNN was tested in relatively simple environments. However, it should be noted, that due to the permutation invariance of the network, a network initialized for any of the environments has no built-in knowledge about what action the activation of any specific output neuron will result in. This must be inferred during the agent's lifetime through interactions with the environment.

The freedom to optimize parameters in any environment potentially enables SFFNs to be candidate models for foundation models for RL-type tasks. This would require training in as wide a range of environments as possible. The more environments the SFNN can generalize across, the more useful it would potentially be in future tasks. However, scaling up the approach to many more environments could face some obstacles.

The experiments in this paper used environments with small input and output sizes. Challenges related to the symmetry dilemma are magnified when input and output sizes are increased. More specifically, more input elements potentially make it harder for the input synapses with shared parameters to distinguish them and make properly individualized synaptic strengths for each of them. More experiments are needed to determine whether more types of neurons and synapses in the hidden reservoir are sufficient to mitigate the symmetry dilemma in high-dimensional environments.
Similarly, future experiments should examine the SFNN approach in environments with continuous action spaces. Since choosing discrete actions, as in the environments used here, forcefully breaks symmetry among output neurons' signals, continuous action spaces might prove more challenging.

Breaking the symmetry of the network by placing neurons in the same layer with a random adjacency matrix requires the training phase to incorporate different permutations of the adjacency matrix and ordering of neurons so that the evolved parameters do not overfit to any specific configuration. Further, looking at the second column in Figure ~\ref{fig:life_progress}, the SFNN shows better average performances in two out of three environments, when a single adjacency matrix was used in all lifetimes. This points towards some adjacency matrices being easier to adapt to than others. With these points in mind, instead of relying on random connectivity future studies could potentially benefit from combining the SFNN approach with an algorithm for structural development \citep{najarro2023towards}.

In future studies an interesting extension to the training and testing of the models' ability to learn multiple different tasks during a single lifetime. How difficult is it for the networks to reconfigure themselves to solve a new task after having already organized from its random initialization to solve one task? To be successful in such task switching, it will likely be necessary to incorporate multiple tasks into the agents' lifetimes during training. \newline

Artificial neural networks, which can be optimized across different environments to quickly adapt their synaptic weights to form a functional network, have the prerequisites to develop into a general learner. Through the promising results demonstrated with the introduced SFNN approach in this paper, we hope to contribute to this direction. With a relatively small number of evolved parameters, an SFNN improves during its lifetime on three different control tasks. We believe that scaling up this and related approaches to many more tasks has the potential to yield new exciting opportunities in the field of artificial agents.

\section*{Acknowledgements}
This work was supported by the European Union (ERC, GROW-AI, 101045094), and a Sapere Aude: DFF Starting Grant (9063-00046B).

\bibliographystyle{ACM-Reference-Format}
\bibliography{refs}

\end{document}